# Autonomous Control of a Line Follower Robot Using a Q-Learning Controller

Sepehr Saadatmand
*Department of Electrical and Computer Engineering*
*Missouri University of S&T*
Rolla, Missouri, USA
sszgz@mst.edu

Sima Azizi
*Department of Electrical and Computer Engineering*
*Missouri University of S&T*
Rolla, Missouri, USA
sacc5@mst.edu

Mohammadamir Kavousi
*Department of Electrical and Computer Engineering*
*University of California, Riverside*
Riverside, California, USA
mkavo003@ucr.edu

Donald Wunsch
*Department of Electrical and Computer Engineering*
*Missouri University of S&T*
Rolla, Missouri, USA
dwunsch@mst.edu

*Abstract*—In this paper, a MIMO simulated annealing (SA)–based Q-learning method is proposed to control a line follower robot. The conventional controller for these types of robots is the proportional (P) controller. Considering the unknown mechanical characteristics of the robot and uncertainties such as friction and slippery surfaces, system modeling and controller designing can be extremely challenging. The mathematical modeling for the robot is presented in this paper, and a simulator is designed based on this model. The basic Q-learning methods are based pure exploitation and the ε-greedy methods, which help exploration, can harm the controller performance after learning completion by exploring nonoptimal actions. The simulated annealing–based Q-learning method tackles this drawback by decreasing the exploration rate when the learning increases. The simulation and experimental results are provided to evaluate the effectiveness of the proposed controller.

## I. INTRODUCTION

In recent years, the penetration of the robotics application is steadily rising thanks to the development of new technologies in mechanical, electrical, and computer engineering. Industrial robots have been functioning in several industries for decades, and state-of-art surgical robots have been used in medical applications [1], [2]. However, designing an advanced complicated industrial robot can be overwhelming for newly graduated engineers. Therefore, robotic competitions are designed to familiarize students and young engineers with the basic concepts of the robotics field of study.

Robotic competitions are being held annually in several countries in different leagues. The best teams in each league can participate in the robotic world cup competition (RoboCup), which is the most well-known competition in the field of robotics. The competition is held in two different categories: (i) university students and (ii) high school students. The most famous leagues are RoboCupSoccer (including middle size, small size, and humanoid), RoboCupRescew, and

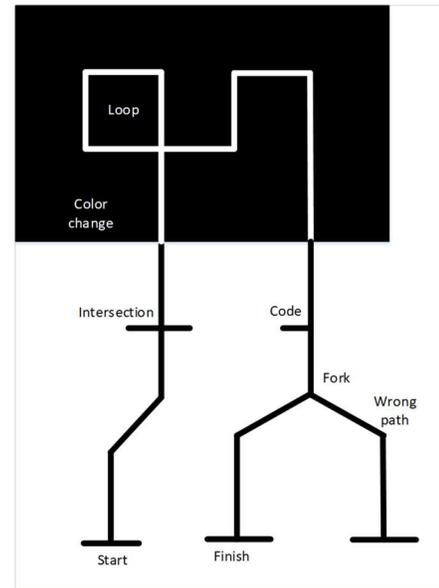

Figure. 1. An example path for a line follower competition

RoboCup@Home. In RoboCupRescew, one important part is to follow a specific line to navigate to the victims and rescue them. Therefore, line following robots have become one of the most popular competition categories in robotic events [3].

The line follower robot league is based on a simple rule of following a specific line trajectory in the shortest amount of time. The typical path is a black line in a white plane, but the rules have been updated several times to make the competition more complicated and more interesting. The updated rules also include: (i) the line color can be white and the plain color can be black, (ii) there can be a loop in the path, (iii) there are intersections, and (iv) there is a possibility of a coded fork. Figure 1 illustrates an example of a path designed for the competition [4].

A typical line follower robot includes four different circuit blocks. The first circuit block is the power block, which includes a supply voltage, (typically a battery), and a motors speed with a pulse width modulation (PWM) signal by applying the controller scheme [5].

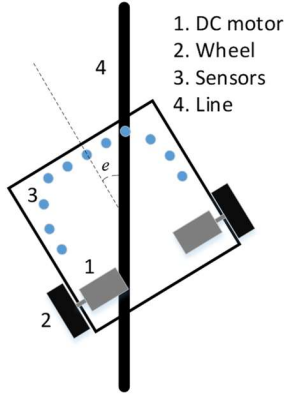

Figure. 2. The error definition based on the robot position on the line

The most common method to control a line follower robot is a proportional controller. The error signal $e$ is defined based on the distance of the sensor that is on the line and the center sensor, as shown in Figure 2. To design a proportional controller, the linear/linearized model of the system is required. Considering all the uncertainties such as frictions and unknown motor parameters, finding the linearized model of the system can be challenging. The most common method to design the proportional controller for a line follower robot is based on trial and error, which can be time consuming due to the procedure of updating weights and reprogramming the microcontroller.

Nowadays, reinforcement learning (RL) Techniques have been used in a large variety of applications such as Atari games, robotic arms, text analysis [6], and power systems [7]-[11]. Reinforcement learning is based on an agent that controls a system. An evaluation system criticizes the effectiveness of the control command and either rewards or penalizes the agent based on its performance [12]. Q-learning method is an RL technique with the objective of cost-to-go function minimization. The exploration-exploitation dilemma have been studied in several publications, and in this paper a new technique is introduced to overcome this drawback.

The main contribution of this paper is to propose a simulated annealing (SA)-based Q-learning technique to control a basic line follower robot. Basic rules have been applied in this paper to follow a simple path without complex traps such as forks.

The rest of this paper is organized as follows. The fixed/variable regulator to provide the logic voltage and the motor drive voltage. The second block is the sensor circuit, which is designed to recognize the path using a photo/infrared-based transmitter and emitter. The infrared system is preferred to minimize the environmental interference. The third block is the motor drive block, which drives the motors using the microcontroller command. The last block is the control block, which includes a microcontroller. The microcontroller receives the sensors' information and locates the path, and controls the mathematical modeling of a line follower robot is presented in Section II. The Q-learning technique is explained in Section III, and the simulation results are provided in Section IV to evaluate the effectiveness of the proposed controller. Lastly, we conclude the paper in Section V.

## II. MATHEMATICAL MODELING FOR THE MOTION

Figure 3 illustrates the robot dimensions and features. In this figure, $a$ and $b$ are the robot width and length, respectively. The wheel radius is shown by $r$, and the center of the mass point is shown by $M$. The angle between the robot and the y axis is shown by $\delta$. At each time step, it is assumed that the center of mass of the robot is located in $(x_0, y_0)$ and the robot angle is $\delta_0$. The objective of the motion equation is to provide the next location for the center of mass and the next robot angle. The motion equations are presented in the discrete region with the sampling time of $T_s$. Two motion scenarios are considered to mathematically model the motion behavior. The first scenario is when both wheels rotate in the same direction. In both scenarios, first the robot is transferred to the origin and then the motion equation is applied. In this scenario, the motion is in two parts: (i) straight movement, and (ii) the rotation around the center of the wheel with less speed. The straight movement equations can be written as

$$w_f = \text{sign}(w_l) \times \min(|w_l|, |w_r|) \quad (1)$$

$$V_f = w_f \times (2\pi r) \quad (2)$$

$$x_f = -\sin\delta \ V_f T_s \quad (3)$$

$$y_f = \cos\delta \ V_f T_s \quad (4)$$

where $w_f$, $V_f$, $x_f$, and $y_f$ are the resultant of the left and right rotational speed, the forward speed, the forward movement in the $x$ axis direction, and the forward movement in the $y$ axis direction, respectively. In addition, the rotating movement can be expressed by

$$w_{rt} = \max(|w_l|, |w_r|) - \min(|w_l|, |w_r|) \quad (5)$$

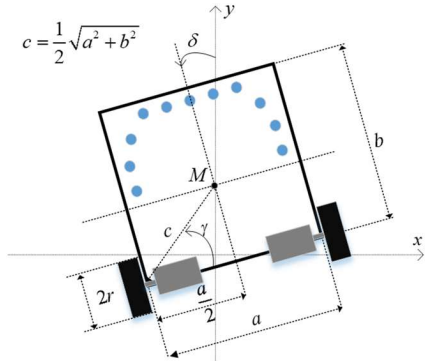

Figure. 3. Robot dimensions and parameters

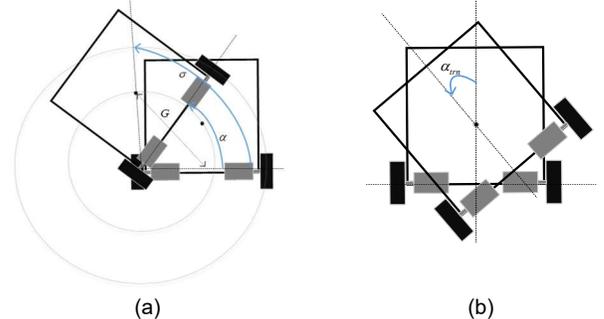

(a)          (b)

Figure. 4. (a) the robot rotation when one motor is on and the other is off (b) the robot turn when both motors rotate with the same speed in the opposite direction

$$\alpha = w_{rt}\left(\frac{r}{a}\right)T_s \tag{6}$$

$$\gamma = \mathrm{atan}(\frac{b}{a}) \tag{7}$$

$$\sigma = \gamma + \alpha \tag{8}$$

$$G = \sqrt{\left(\frac{a}{2}\right)^2 + c^2 - a \cdot c \cdot \cos\sigma} \tag{9}$$

$$x_{RotOrg} = G \cdot \sin(\alpha) \tag{10}$$

$$y_{RotOrg} = G \cdot \cos(\alpha) - 0.5 \cdot b \tag{11}$$

$$[x_{Rot}, y_{Rot}] = O(x_{RotOrg}, y_{RotOrg}, \delta) \tag{12}$$

where $w_{rt}$, $x_{RotOrg}$, $y_{RotOrg}$, $x_{Rot}$, $y_{Rot}$, and $O(\cdot)$ are the rotational speed, the movement in the x axis and the y axis in their original axis, the equal x and y movement in the transferred origin, and the origin transfer function, respectively. Figure 4 illustrates the defined angles and movements. By adding the movement caused by the forward motion and rotational motion, the new location for the center of mass and the new robot angle can be computed as

$$x_{out} = x_f + x_{Rot} + x_0 \tag{13}$$

$$y_{out} = y_f + y_{Rot} + y_0 \tag{14}$$

$$\delta_{out} = \delta_0 + \alpha \tag{15}$$

where $x_0$, $y_0$, and $\delta_0$ are the last location and angle of the robot, and $x_{out}$, $y_{out}$, and $\delta_{out}$ are the new location and angle of the robot, respectively. Equations (1)-(15) explain that the condition of both motor speeds are positive. For the scenario where both speeds are negative, the equation can be easily written based on (1)-(15). For the second scenario where the motors' rotational speeds are opposite, the motion equation can be written as

$$w_{rt} = w_r + w_l \tag{16}$$

$$w_{trn} = -\mathrm{sign}(w_r) \cdot \min(w_r, w_l) \tag{17}$$

$$\alpha_{trn} = \left(\frac{2r}{a}\right) \cdot w_{trn} \cdot T_s \tag{18}$$

$$x_{out} = x_{Rot} + x_0 \tag{19}$$

$$y_{out} = y_{Rot} + y_0 \tag{20}$$

$$\delta_{out} = \alpha + \alpha_{trn} + \delta_0 \tag{21}$$

where $w_{rt}$ is the rotational speed, and its movement that can be computed using (5)-(12). Equation (17) explains the rotational turning speed that happens when $w_r = -w_l$. The equation for the change in angle regarding to the turning is explained via (21).

---

**Algorithm 1: The Q-learning one step technique**

1. Initialize the $Q(s, a)$ arbitrarily
2. For each epoch repeat:
   - I. Chose a random initial state
   - II. For each step in this episode repeat:
     - i. Select the action regarding to the policy
     - ii. Implement the selected action, compute the rewards and observe the next state
     - iii. $Q(s_t, a_t) \leftarrow Q(s_t, a_t) + \alpha \left(r_t + \gamma \max_a(s_{t+1}, a) - Q(s_t, a_t)\right)$
     - iv. $s \leftarrow s'$

   Continue until $s$ is not in the state domain or the episode ends.

---

## III. SIMULATED ANNEALING Q-LEARNING

In this section the Q-learning technique based on simulated annealing (SA) is explained. Firstly, the basic Q-learning is explained and its drawbacks are defined, and then it will be clarified how the SA algorithm can help to overcome basic Q-learning concerns.

### A. Q-Learning Algorithm

Machine learning techniques have been used in various applications [13]. In 1989, Watkins presented the Q-learning algorithm as one of the most highlighted algorithms in the field of reinforcement learning [14]. Temporal difference (TD) learning is one the most well-known RL techniques, and Q-learning is categorized as a special case of TD learning by some researchers.

The basic rule of the Q-learning algorithm can be expressed as

$$Q(s_t, a_t) \leftarrow Q(s_t, a_t) + \alpha \left(r_t + \gamma \max_a(s_{t+1}, a) - Q(s_t, a_t)\right) \tag{22}$$

where $s_t$, $a_t$, $r_t$, and are the state, action, and the rewards at time $t$, respectively. In addition, $\gamma$, $\alpha$, and $Q(s_t, a_t)$ are the discount factor $\gamma \in [0,1]$ to guarantee the divergence of the value function, the learning rate, and the value function at time $t$ with state $s_t$, and the chosen action of $a_t$, respectively.

The pure exploitation approach is used in the original Q-learning. In other words, to select the action, only the optimal policy is being followed. However, this method can be inefficient when it gets stocked in local minima. To make sure there is sufficient exploration in learning, the agent needs to be allowed to select a nonoptimal action. As a case in point, to overcome the exploration concern in ε-greedy methods, the action can be chosen as a nonoptimal with the fixed probability of ε. Although this approach helps the more accurate and the exploration can decreases exploration, by increasing the learning process, the policy becomes the effectiveness of the controller. In other words, the exploration needs to decrease by increasing the learning process.

### B. SA-based Q-learning

The simulation annealing technique is one of the most common optimization methods that mimics the behavior of the steal annealing process [15]. The transition procedure from state $a$ to state $b$ is based on its probabilities, which can be explained as

$$P(a \to b) = \begin{cases} 1, & \text{if } f(b) \geq f(a) \\ e^{\left(\frac{f(b)-f(a)}{T}\right)}, & otherwise \end{cases} \tag{23}$$

where $a$ is the current state, $b$ is the next state, $T$ is the synthetic temperature, and $f(\cdot)$ is the value of the optimization cost function. In another words, SA guarantees that if the next action state is not optimal, and there is still a probability to get into the nonoptimal state and by decreasing the temperature thus increasing the annealing process, this probability is reduced. By implementing the idea of the SA in the basic Q-learning algorithm we can tackle the exploration-exploitation dilemma.

Contrary to the basic Q-learning, selecting the action in SA-based Q-learning is not only based on the current optimal policy, but the nonoptimal action also has the chance to be chosen. In other words, it is not only based on pure exploitation,

| Algorithm 2: The SA-based Q-learning one step technique |
|---|
| 1. Initialize the $Q(s,a)$ arbitrarily |
| 2. For each epoch repeat: |
|     I. Chose a random initial state |
|     II. For each step in this episode repeat: |
|         i. Select the optimal action regarding to the policy as $a_o$ |
|         ii. Select a random action as $a_r$ |
|         iii. Generate a random number as $\sigma \in [0,1]$ |
|         iv. If $\sigma < e^{\left(\frac{Q(s_t,a_r)-Q(s_t,a_o)}{T}\right)}$: then $a_t \leftarrow a_r$ |
|         v. Else: $a_t \leftarrow a_o$ |
|         vi. Implement the action $a_t$, compute reward $r_t$, and observe the next state |
|         vii. $Q(s_t,a_t) \leftarrow Q(s_t,a_t) + \alpha \left(r_t + \gamma \max_a(s_{t+1},a) - Q(s_t,a_t)\right)$ |
|         viii. $s \leftarrow s'$ |
|     III. Recalculate $T$ |
| Continue until $s$ is not in the state domain or the episode ends. |

but the agent can explore throughout different action options. In this method, a parameter (like the temperature in the SA) defines the portion of exploration and exploitation at each time step.

The temperature decreasing process can be arbitrary or it can follow any dropping pattern, but in this paper the temperature can be computed as follows

$$T = \frac{1}{\beta \cdot t} \quad (24)$$

where $\beta$ is a positive constant number.

When comparing Algorithm 1 and Algorithm 2, there are only two main differences: (i) the selection of random actions and (ii) the computation of the temperature. Therefore, the implementation of SA-based Q-learning does make the basic Q-learning technique more complex.

## IV. SIMULATION RESULTS

To simulate the proposed controller technique, the parameters and expression regarding the SA-based Q-learning algorithm need to be explained. In this paper, the robot error can be defined as the angle between the sensor on the line, the point of center of mass, and the central sensor, as shown in Figure 2. The reward function is defined as follows

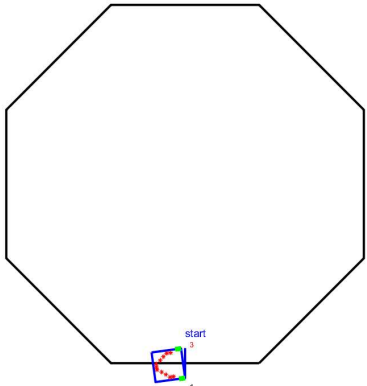

Figure. 5. Robot dimensions and parameters

Table I. Robot and controller parameters

| Robot Dimensions and Characteristics | | |
|---|---|---|
| Feature | Notation | Value |
| Length | $a$ | 20 cm |
| Width | $b$ | 25cm |
| Motor speed | $w_m$ | 600 RPM |
| Wheel radius | $r$ | 2.5 cm |
| Learning rate | $\alpha$ | 0.01 |
| Discount factor | $\gamma$ | 0.99 |

$$r_t = \Delta e + \left(\frac{w_l+w_r}{2*w_m}\right) - \Delta t \cdot e. \quad (25)$$

The first part of the reward function in (25) introduces the reward regarding the error corrections; the second part encourages the controller to go as fast as possible; and the last part is to guarantee that the robot recovers from the error as fast as possible. The speed of the right and the left motor are the control variables of the robot. In order to define the action set, two different scenarios are considered in this paper. In the first scenario, it is assumed that there is only one control, which can be chosen from the following data set

$$a = [y_l, y_r], \quad y_l, y_r \in \{-1,0,1\}, \quad (26)$$

where $y_l, y_r$ are the portion of the maximum speed for the right and left motor, respectively. In the second scenario, to smoothen the motion, $y_l, y_r$ can be chosen from a set of data with more options as $\{-1, -0.9, -0.8, \ldots, 0.8, 0.9, 1\}_{21}$. To generate the action set of data like the first scenario, the set of action includes 441 ($21 \times 21$) different options, which increases the learning procedure. Therefore, two control variables are considered for this scenario, show as follows

$$\begin{aligned} a_l &= y_l, & y_l &\in \{-1, -0.9, -0.8, \ldots, 1\}, \\ a_r &= y_r, & y_r &\in \{-1, -0.9, -0.8, \ldots, 1\}, \end{aligned} \quad (27)$$

where $a_l$ and $a_r$ are the action selection for the left motor and the right motor, respectively. Figure 5 illustrates the simulation environment. As shown, the robot is on a designed path with no trap on it, and the goal is to start the path and follow it in the right direction until it reaches the starting point within the shortest amount of time. Since there is no loop in the path to change the path rotation, the direction of the path is also chosen randomly at the starting point, which means that the path can be either clockwise or counter-clockwise. The robot parameters are shown in Table I, which follows the exact parameters for the experimental robot.

The final score is a combination of the followed path and the time and can be computed as

$$S = (10 \times n) - \frac{t_{rec}}{n} \quad (28)$$

where $n$ is the number of completed line segment, and $t_{rec}$ is the time period from starting to the end of the episode. The second part in the right hand side of (28) illustrates that the slower the robot goes, the more penalties it receives.

Figure 6 shows the final score for the robot in different episodes. In this figure, the performance of a P-controlled, $\varepsilon$-greedy Q-learning MISO, SA-Q-learning MISO, and SA-Q-learning MIMO is illustrated. The x and y axis show the episode number and the episode score, respectively. Each episode starts from the starting point and ends when one of the following conditions happens: (i) the robot completes the path and reaches the end point, (ii) the robot loses the track, and (iii) the robot turns and follows the track in the opposite direction. As expected, the traditional P-controlled robot performs similarly in all episodes; however, the performance of the Q-learning–based robots improves by proceeding through the learning process. As the simulation results shows the performance of both the ε-greedy and the SA-based Q-learning controller improves by iteration until it is trained. After learning the process, the exploration in ε-greedy continues and causes nonoptimal scores, on the contrary the SA-based technique reduces the exploration by the increase in learning. As expected, the limitation in control reduces the effectiveness of the MISO with only five modes of control. Nonetheless, the MIMO controller, which has 421 control modes, performs better than MISO five-mode learning. The learning process for MIMO controller is longer compared to the MISO controller; however, after sufficient training, the performance is better than all the simulated controllers. Figure 6 (b) illustrates the final episodes, when the agent is completely trained. As the score shows, all controller scores are between 70 and 80, which proves that all of them are able to get to the end point; however, the ε-greedy-based tends to have nonoptimal results since it does not stop the exploration. The small difference is because of the time penalty, and the MIMO SA-Q-learning with more accurate weights is able to follow the line more smoothly and more quickly.

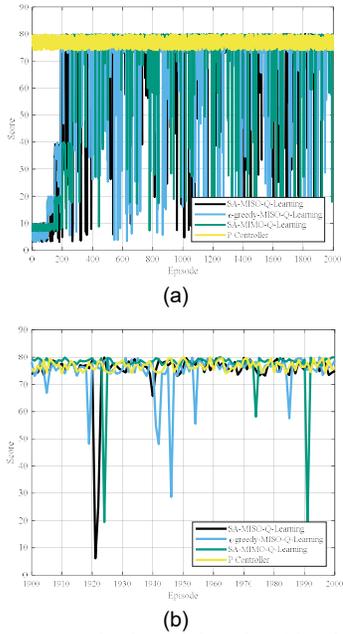

Figure. 6. Performance evaluation of the robot (a) episodes 1-2000, (b) episodes 1900-2000

Table II. Simulation results for the robot scores on the complex path

| The best score out of five tries | |
|---|---|
| Algorithm | Score |
| P controlled | 317.72 |
| ε-greedy Q-learning single controlled | 314.36 |
| SA-based Q-learning single controlled | 314.53 |
| SA-based Q-learning double controlled | 318.26 |

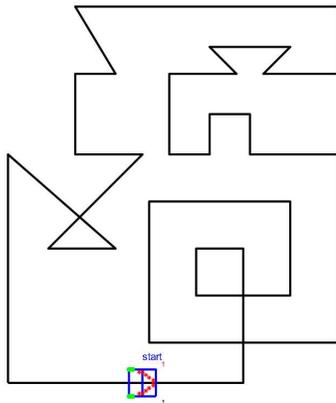

Figure. 7. Complicated line path to test the trained algorithms

To evaluate the performance of the proposed techniques, the trained algorithm has been tested on a more complex path, as shown in Figure 7. Table II illustrates the final results for all the controllers. As expected, the MIMO SA-based Q-learning controller performs more efficiently compared to the rest of the controllers.

## V. EXPERIMENTAL RESULT

Figure 8 illustrates the robot in top/bottom view and describes its important parts. In order to test the algorithm, the experimental robot is built based on the ATMEGA64 microcontroller. To find the path, 32 infrared transmitters and receivers are used. To convert the analog output to a digital output buffer ICs, specifically SN74LS245N, are used. The digitized output of the sensor is read directly by the microcontroller input pins. Two gearbox motors with the nominal RPM of 600 are chosen to move the robot. All four control techniques are implemented and tested on the robot. The experimental results at each epoch are shown in Figure 9. As expected from the simulation, the experimental results verify that MIMO SA-based Q-learning control is the most effective control technique; however, to achieve the best performance, the proposed algorithm needs to be trained completely. As shown in Figure 9 (b), even after the training process, the final score for all the episodes is not optimal because while of the simulation, in a real environment there are uncertainties such as slippery surfaces and motor noises, which can cause nonoptimal results. After training the reinforcement learning–based controller, the performance of the robot on a complicated line map is tested. Table III shows the best score out of five tries for each controller. The robot performance on a complicated path is shown in Figure 10.

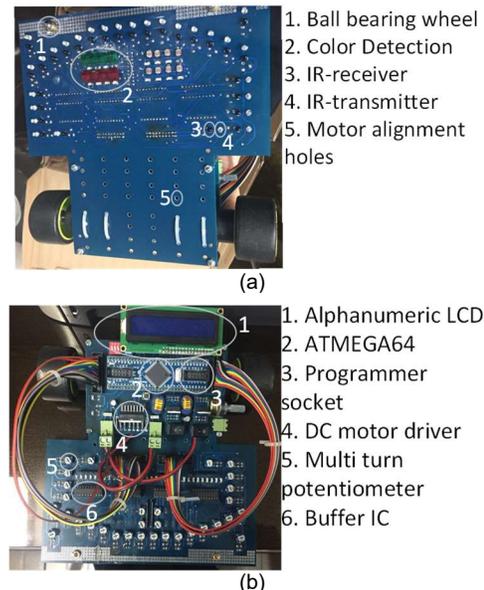

Figure. 8. (a) The bottom view of robot including sensor board, (b) the top view including control board

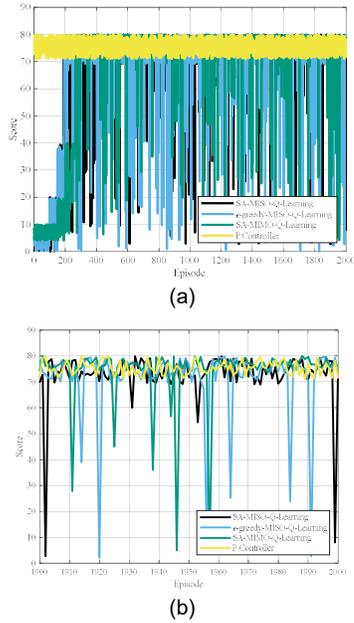

Figure. 9. The robot score in each episode

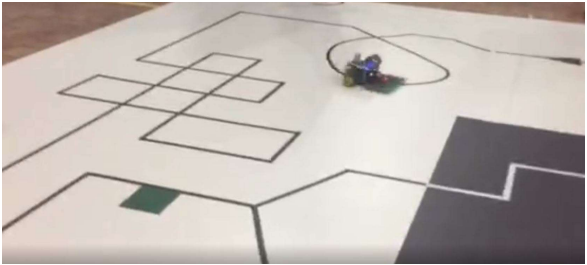

Figure. 10. Experimental robot test on a complex line map

Table III. Experimental results for the robot scores on the complex path

| The best score out of five tries ||
|---|---|
| Algorithm | Score |
| P controlled | 516.32 |
| ε-greedy Q-learning MISO | 432.93 |
| SA-based Q-learning MISO | 433.34 |
| SA-based Q-learning MIMO | 518.95 |

## VI. Conclusion

The importance of robotics has been rapidly increasing for decades. Robotic competitions have emerged to prepare students to work and design complicated advanced robots. One of the simplest types of these robots is the line follower robot. According to the author's knowledge, there has been no research on mathematical modeling and controller designing for line follower robots. In this paper, a thorough mathematical model is proposed. The most common controller for the line follower robots is the proportional controller. To design a P-controller, the exact parameters of the system need to be known, which is a challenging task. Therefore, trial and error is the most common technique to tune the controller parameters, which is not optimal. This paper presents a SA-based Q-learning controller to optimally control the robot. The simulation and experimental results show the effectiveness of the MIMO SA-base Q-learning. Moreover, a comparison between the proposed method and three different methods, including (i) MISO SA-based Q-learning, (ii) MISO ε-greedy Q-learning, and (iii) proportional controller are provided to clarify the advantages of the proposed method.